\documentclass{arxivpubl}
\usepackage{eg2023}
 
\ConferencePaper        
\usepackage[T1]{fontenc}
\usepackage{dfadobe}
\usepackage[table]{xcolor}
\usepackage{booktabs} 
 
\usepackage[ruled]{algorithm2e} 

\SetAlFnt{\small}
\SetAlCapFnt{\small}
\SetAlCapNameFnt{\small}
\SetAlCapHSkip{0pt}

\biberVersion
\BibtexOrBiblatex
\usepackage[backend=biber,bibstyle=EG,citestyle=alphabetic,backref=true]{biblatex} 
\electronicVersion
\PrintedOrElectronic
\ifpdf \usepackage[pdftex]{graphicx} \pdfcompresslevel=9
\else \usepackage[dvips]{graphicx} \fi

\usepackage{egweblnk} 
\usepackage{amsfonts}
\usepackage{array}
\usepackage{booktabs}
\usepackage{enumitem}
\usepackage{float}
\usepackage{mathtools}
\usepackage{multirow}
\usepackage{setspace}
\usepackage{subcaption}
\usepackage{xfrac}
\usepackage{pifont}
\usepackage{amsmath}
\usepackage{xcolor}
\usepackage{color, colortbl}

\DeclareUnicodeCharacter{3000}{  }
\DeclareUnicodeCharacter{2212}{-}
\newcommand{\cmark}{\textcolor{green!80!black}{\ding{51}}}
\newcommand{\xmark}{\textcolor{red}{\ding{55}}}

\newcolumntype{L}[1]{>{\raggedright\let\newline\\\arraybackslash\hspace{0pt}}m{#1}}
\newcolumntype{C}[1]{>{\centering\let\newline\\\arraybackslash\hspace{0pt}}m{#1}}
\newcolumntype{R}[1]{>{\raggedleft\let\newline\\\arraybackslash\hspace{0pt}}m{#1}}

\setlist[itemize]{noitemsep, topsep=0pt}
\setlist[enumerate]{noitemsep, topsep=0pt}

\newcommand\numberthis{\addtocounter{equation}{1}\tag{\theequation}}

\newcommand{\parens}[1]{\left(#1\right)}
\newcommand{\braces}[1]{\left\{#1\right\}}
\newcommand{\bracks}[1]{\left[#1\right]}

\newcommand{\norm}[1]{\left\Vert#1\right\Vert}

\definecolor{Gray}{gray}{0.9}
\definecolor{RDcolor}{rgb}{0.0, 0.8, 0.75}

\definecolor{AGcolor}{rgb}{0.8, 0.0, 0.0}

\definecolor{darkorange}{rgb}{1.0, 0.55, 0.0}

\title{IMoS: Intent-Driven Full-Body Motion Synthesis for Human-Object Interactions}

\begin{document}

\author[A. Ghosh, R. Dabral, V. Golyanik, C. Theobalt, P. Slusallek]
{\parbox{\textwidth}{\centering Anindita Ghosh$^{1,2,3}$
Rishabh Dabral$^{2,3}$
Vladislav Golyanik$^{2,3}$
Christian Theobalt$^{2,3}$
Philipp Slusallek$^{1,3}$ 
    } \\
    {\centering $^1$ German Research Center for Artificial Intelligence (DFKI)}
    \\
{\centering $^2$Max-Planck Institute for Informatics (MPII)}
\\
{\centering $^3$Saarland Informatics Campus}
\\
{\centering{ \texttt{\href{https://vcai.mpi-inf.mpg.de/projects/IMoS/}{https://vcai.mpi-inf.mpg.de/projects/IMoS}}} }
}
\teaser{
    \centering
    \includegraphics[width=0.9\linewidth]{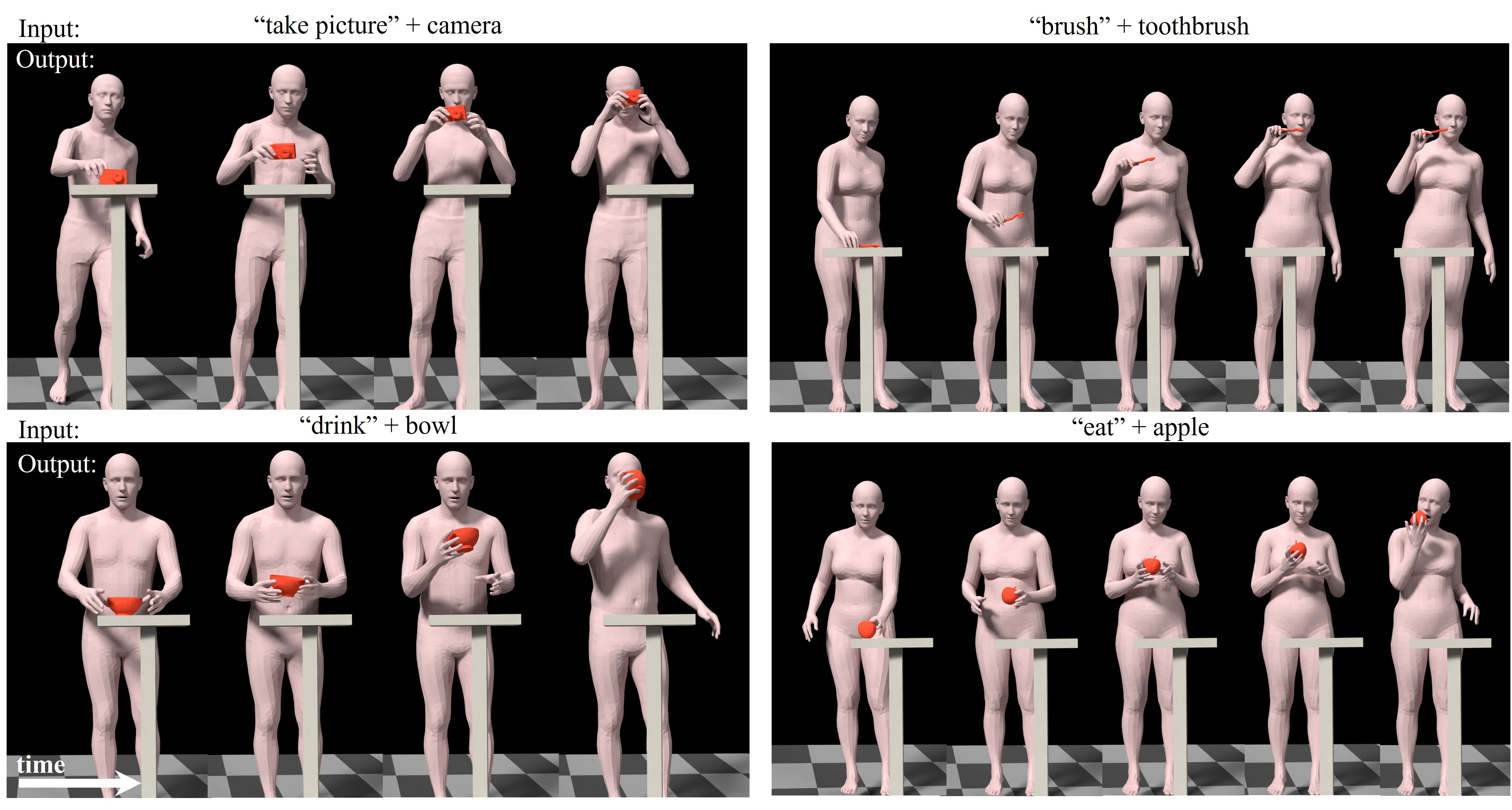}
    \caption{
    \textbf{Visualizations of Motion Sequences of Virtual Characters Performing Various Intended Actions with Different Objects, as Generated by Our Method.}
    We synthesize the full-body pose sequences and the 3D object positions from text-based instruction labels. 
    Our method can synthesize both single-handed and two-handed interactions depending on the intent and the type of object used. 
    }
    \label{fig:teaser}
}

\maketitle

\begin{abstract}
Can we make virtual characters in a scene interact with their surrounding objects through simple instructions? 
Is it possible to synthesize such motion plausibly with a diverse set of objects and instructions? 
Inspired by these questions, we present the first framework to synthesize the full-body motion of virtual human characters performing specified actions with 3D objects placed within their reach.
Our system takes textual instructions specifying the objects and the associated `intentions' of the virtual characters as input and outputs diverse sequences of full-body motions.
This contrasts existing works, where full-body action synthesis methods generally do not consider object interactions, and human-object interaction methods focus mainly on synthesizing hand or finger movements for grasping objects.
We accomplish our objective by designing an intent-driven full-body motion generator, which uses a pair of decoupled conditional variational auto-regressors to learn the motion of the body parts in an autoregressive manner.
We also optimize the 6-DoF pose of the objects such that they plausibly fit within the hands of the synthesized characters.
We compare our proposed method with the existing methods of motion synthesis and establish a new and stronger state-of-the-art for the task of intent-driven motion synthesis.
\end{abstract}

\section{Introduction}\label{sec:intro}
Humans regularly use and interact with objects in numerous ways in the real world. 
Interactions like eating a fruit or brushing the teeth, as shown in Fig.~\ref{fig:teaser}, are part of our daily routines.
Being able to synthesize such interactions in a \textit{virtual} 3D environment through textual instructions has widespread applications in several areas, including computer graphics and robotics~\cite{ahuja2020style, intercap_gcpr2022, wang2022reconstruction}, movie script visualization~\cite{hanser2009scenemaker} and game design~\cite{game_design}.
For instance, in a digitally created movie scene or a virtual role-playing game, it is natural for the character to interact with the scene objects based on a set of instructions, such as yielding tools, using objects, or eating various items.
Manually modeling such 3D character-object interactions or \textit{intentions} is time-consuming and laborious, 
when we desire to synthesize a variety of possible motions with the same intention and object.

In this context, many recent methods automatically synthesize motions for virtual characters by encoding control signals such as music~\cite{Dancing_2_music, li2020learning, AutoDance2021}, speech~\cite{bhattacharya2021speech2affectivegestures, habibie, gesturematching22} or text, either as sentences ~\cite{guo2020action2motion, bhattacharya2021text2gestures, Ghosh_2021_ICCV, petrovich22temos} or as high-level action descriptions~\cite{text2action, lin2018generating, l2p}.
Methods synthesizing full-body pose sequences typically follow an autoregressive approach to maintain continuity in the synthesized motions~\cite{2020-TOG-MVAE, humor2021, Guo_2022_CVPR}. 
These autoregressive motion synthesis frameworks predict short-term future sequences from a short history.
There are also several methods for hand-object interactions~\cite{GraspingField:3DV:2020, GRAB:2020, jiang2021hand, manipnet, christen2022dgrasp}, which focus on generating only the wrist and finger movements for grasping various objects.
However, modeling hand motion alone is insufficient to create a plausible motion sequence for an intent-driven virtual character. 
Instead, \textit{we believe it is crucial to operate in the space of full-body motion synthesis.} 
There are two prime reasons for this.
\textit{Firstly}, synthesizing full-body movements allows for a broader range of interactions (Fig.~\ref{fig:teaser}).
For several intents, such as eating, drinking, inspecting, passing, or exchanging objects between hands, the head, the arms, and the torso are also part of the complete action sequence~\cite{GRAB:2020}.
\textit{Secondly}, trivially attaching the synthesized hand motion to the remaining body~\cite{Puig_2018_CVPR} leads to an uncanny and physically implausible motion generation (see suppl. video).
Further, recent works~\cite{taheri2021goal, saga} have demonstrated the ability to generate whole-body grasping motion starting from a T-Pose \textit{till} the moment of the grasp.
However, synthesizing a plausible motion sequence \textit{after} the first grasp moment, based on an intent guiding the human-object interaction, remains unaddressed.
\begin{figure}[t]
\centering
    \includegraphics[width=0.8\columnwidth]{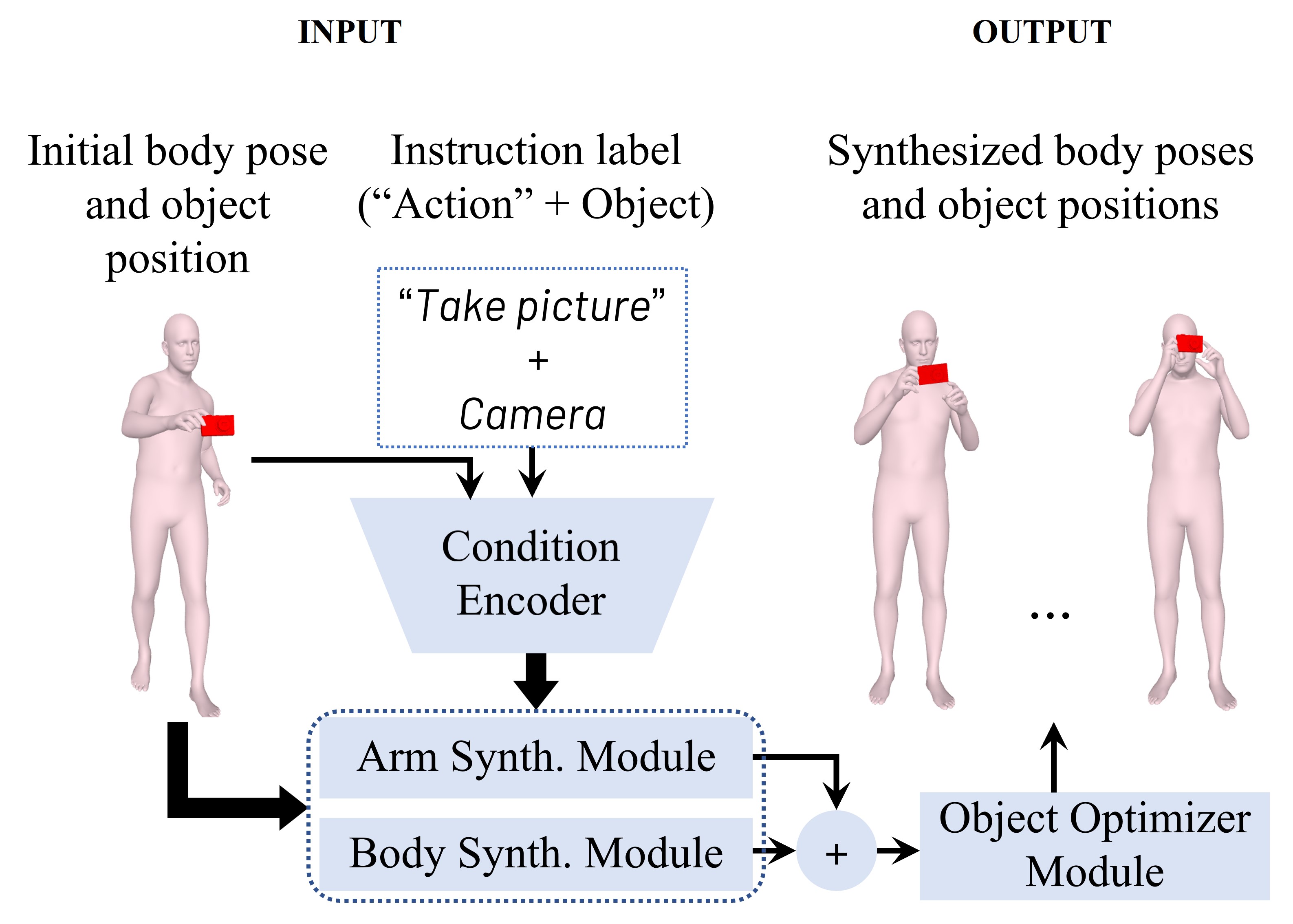}
    \caption{\textbf{Overview of Our Intent-Driven Full-Body Motion Generator.} Our model takes in the initial 3D body poses and object positions (upper-left) and instruction labels (upper-middle) describing the object types and the intended actions.
    We design a pair of decoupled conditional variational auto-regressors, the Arms Synthesis Module and the Body Synthesis Module (lower-middle), to separately synthesize the arms and the rest of the body. We also design a Condition Encoder (middle) to condition our decoupled autoregressors based on the input instruction labels and the body shape parameters. We concatenate our synthesized arm and body motions and use our Object Optimizer Module (lower-right) to optimize the $6$-DoF parameters of the object while satisfying the grasping constraints. Our model outputs the synthesized full-body motion sequence together with the object positions (upper-right).}
    \label{fig:overview_fig}
\end{figure}

 To address these limitations, we propose IMoS, a novel framework to synthesize diverse, full-body motion sequences of human-object interactions.
 Crucially, we synthesize the motions based on the input textual instructions consisting of actions (intentions) and objects (Fig.~\ref{fig:overview_fig}). 
We learn generalizable intent encodings from the input intent-object pairs using a CLIP encoder~\cite{clip}, which is a large-scale language model trained on a large corpus of text-image pairs. 
Given the initial body poses and the 3D object positions, we design an intent-driven full-body motion generator model to autoregressively generate full-body motions (Sec.\ref{sec:Intent-Driven Full-Body Motion Generator}).
We follow a decoupling approach and model the arms and the body motions using separate Conditional Variational Autoregressors to make our output arm and body movements more precise.
Since these autoregressors are variational in nature, they allow us to sample diverse motions from the latent space at inference time.  
We also observe that regressing the motion from a larger past context is crucial in modeling long-term temporal dependence between the joints. 
We use a position-encoded self-attention mapping to model correlations between the different joints to allow a broader range of interactions.
Lastly, we perform an optimization routine to estimate the corresponding $6$-DoF object positions relative to the hand position in each frame (Sec~\ref{sec:optimization}).
We use the recovered object positions to condition future motion synthesis.

We train and evaluate our method on the recent GRAB dataset \cite{GRAB:2020} (Sec.~\ref{sec:dataset}), consisting of ${\sim}1.3$K sequences of human-object interactions exhibiting multiple intents.
We quantitatively evaluate our synthesized sequences on established metrics, such as the mean per-joint position error, the average variance error, the Fr\'echet Inception distance, recognition accuracy, diversity, and multimodality, to test the effectiveness of the model.
Further, we conduct a visual perceptual study for subjective evaluation of our synthesized motions compared to recent conditional motion synthesis methods (Sec.~\ref{perceptual_study}).

In summary, our primary \textbf{technical contributions} are threefold: 

\begin{itemize}[leftmargin=*]
    \item A new framework for generating diverse motion sequences of virtual human characters interacting with objects of known shapes placed within their reach, according to text-based instruction labels.
    In contrast to previous works on character-object interactions, our proposed method also optimizes the $6$-DoF object positions in 3D.
    \item  Synthesizing interactions involving \textit{both} hands, 
    including sequences where the character exchanges an object between the hands (``offhand'') -- a previously unexplored setting.
    
    \item Learning separate variational latent embeddings for the arms from the rest of the body to enable diversity in the synthesized motions and accurate synthesis of both-handed interactions.
   
\end{itemize}
\begin{table}[t]
\center  
 \scalebox{0.9}{ 
 \begin{tabular}{ l c  c c c}\toprule 
 \multirow{3}{*}{Method} & \multicolumn{4}{c}{ Motion Synthesis} \\
 				  \cmidrule(lr){2-5} 
 & \shortstack{Full\\Body} &\shortstack{Intent-\\Driven} &\shortstack{Only Till\\ Grasp }  &\shortstack{Object\\ Manipulation}\\
 \addlinespace[2pt] \midrule
\rowcolor{Gray}
GRABNet~\cite{GRAB:2020} & \xmark &  \xmark &\xmark &  \xmark \\  
D-Grasp~\cite{christen2022dgrasp}	& \xmark  &  \xmark& \xmark & \cmark\\  
\rowcolor{Gray}
A2M~\cite{petrovich21actor} & \cmark & \cmark & \xmark & \xmark \\
ACTOR~\cite{petrovich21actor} & \cmark & \cmark & \xmark & \xmark \\
\rowcolor{Gray}
GOAL~\cite{taheri2021goal}& \cmark  &  \xmark&\cmark & \xmark\\  
SAGA~\cite{saga}&  \cmark &  \xmark & \cmark & \xmark  \\ 
\rowcolor{Gray}
IMoS (ours)~ & \cmark &  \cmark &\xmark & \cmark \\ \bottomrule 
\end{tabular}
} 
 \caption{\label{tab:in_out_com}  
\textbf{Overview of the Problem Definitions of Existing Methods.} Our method is the only one combining three important characteristics and the first one to synthesize intent-driven full-body pose sequences for motions with object manipulation.
}
\end{table}

\section{Related Work}
Our work aligns with past works on modeling 3D human-object interactions. We study these works from four vantage points: human pose forecasting and synthesis, human-object 3D interaction modeling, hand-object grasp synthesis, and full-body grasp synthesis.

\paragraph*{Human Pose Forecasting and Synthesis.}
Human pose forecasting methods predict future motions from a sequence of past poses as joint positions~\cite{martinez2017human} or joint rotations~\cite{pavllo2018quaternet, humor2021}.
Recent works on 3D human pose forecasting are stochastic methods~\cite{yuan2020dlow, liu2021aggregated} that use VAEs~\cite{Variational_bayes} or GANs~\cite{goodfellow2020generative} to introduce variability in the output motion sequences.
HuMoR~\cite{humor2021} proposes a CVAE architecture that learns a distribution of pose transitions in the latent space and ensures physical plausibility through post-processing optimization.
Motion-VAE~\cite{2020-TOG-MVAE} learns to drive a character based on a goal position by decoding from a variational latent space. 
Characteristic 3D pose~\cite{diller2022charposes} stochastically predicts future 3D characteristic poses given short sequences of observations. Other human motion synthesis methods are trained to synthesize a motion sequence conditioned on semantic action labels ~\cite{guo2020action2motion, petrovich21actor, diller2022charposes}, or text sentences~\cite{Guo_2022_CVPR, petrovich22temos}. 
Action2Motion~\cite{guo2020action2motion} inputs an action label to generate the human pose in an autoregressive manner using a VAE-GRU. 
Differently, ACTOR~\cite{petrovich21actor} employs a VAE-Transformer to generate the full sequence in one shot.
TEMOS~\cite{petrovich22temos} uses the VAE-Transformer concept on a multi-modal setting to generate motions from text sentences.
Our work extends full-body motion synthesis conditioned on semantic labels by additionally incorporating object interactions.

\paragraph*{Human-Object 3D Interaction Modeling.}
With the availability of several human-object 3D datasets like~\cite{neural_state_machines}, BEHAVE~\cite{bhatnagar22behave}, 
PROX~\cite{PROX:2019},
D3D-HOI~\cite{xu2021d3dhoi}, 
H2O~\cite{Kwon_2021_ICCV_h2o}, GraviCap~\cite{GraviCap2021}, 
joint human-object motion modeling has been an actively researched topic.
Among more recent methods, PHOSA~\cite{zhang2020phosa} reconstructs the human and the object in the scene by jointly optimizing for the reprojection error of the object's silhouette and the human. 
Neural State Machines~\cite{neural_state_machines} synthesize human motion while interacting with objects like chairs or a wall in the scene. 
Likewise, SAMP~\cite{hassan_samp_2021} incorporates a path planning module to improve the character's motion in the scene.
COUCH~\cite{zhang2022couch} synthesizes sitting interactions with couches by generating contact points and then using them to constrain the sitting motion.
All the previous methods perform interactions with large objects (chairs and couches) and are driven by low-level character control.
In contrast, we synthesize fine-grained motions with \textit{handheld} objects using \textit{instruction labels} as input.

\paragraph*{Hand-Object Grasp Synthesis.}
Grasp synthesis has been extensively studied in computer graphics~\cite{elkoura2003handrix, li2007data, kim2015physics, GraspingField:3DV:2020, manipnet} and robotics~\cite{borst2005realistic, hsiao2006imitation, detry2010refining, antotsiou2018task, liu2019generating}.
Analytical approaches have formulated grasp synthesis as a constrained optimization problem satisfying the grasp properties~\cite{krug2010efficient, 6225086}.
Data-driven approaches~\cite{redmon2015real, pinto2016supersizing} focus on learning the representations for synthesizing grasps through machine learning methods. 
More recent approaches~\cite{brahmbhatt2019contactgrasp, GraspingField:3DV:2020, GRAB:2020, jiang2021hand} predict the hand parameters of the MANO hand model~\cite{MANO:SIGGRAPHASIA:2017} for synthesizing a grasp using neural networks.
Many image datasets~\cite{hasson19_obman, zhang2019end, Brahmbhatt_2020_ECCV, Lin_2021_WACV, zhang2021hand} featuring hand-object interaction with contact maps are also currently available.
Taheri et al.~\cite{GRAB:2020} further introduce the GRAB dataset, which captures the contact map from hands and the full-body motions before and during the grasp.
They also propose GrabNet, a network that estimates MANO parameters at the moment of grasp for unseen objects in a coarse-to-fine manner.
\cite{GraspingField:3DV:2020} proposes Grasping Field, a method that learns an implicit representation of the hand-object interaction using a generative model.
Grady et al.~\cite{Grady_2021_CVPR} derive physically plausible hand pose estimation by optimizing estimated hand meshes with contact prediction.
We differ from all these methods as our work focuses on synthesizing \textit{full-body} sequences. 
While modeling hand-object interaction is a well-researched problem, it is inherently limited in its ability to model several types of human-object interactions that require the full human body (\textit{e.g.}, tilting back the head when drinking from a glass).

\begin{figure*}[t]
\centering
    \includegraphics[width=0.9\linewidth]{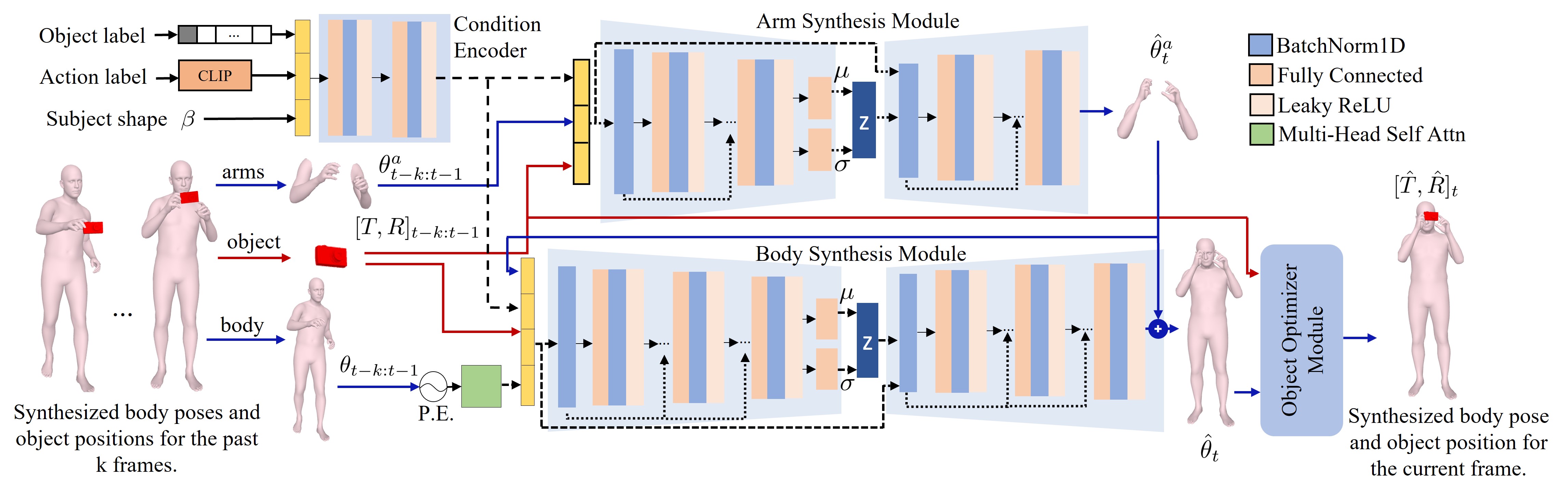}
    \caption{\textbf{Architecture of Our Intent-Driven Full-Body Motion Generator Model.} 
    Given previous $k$ frames of body poses and object positions, we train the arms and the rest of the body separately using our Arm Synthesis (upper-middle) and the Body Synthesis (lower-middle) Modules, respectively.
    We jointly synthesize the entire motion sequences autoregressively, conditioned on the input intent, the object, and the body shape, all encoded through our Condition Encoder (upper-left).
    We use position-encoded self-attention on the past $k$ frames for the body joints before passing them through our Body Synthesis Module. 
    After generating the body pose, our Object Optimizer Module (lower-right) optimizes for the $6$-DoF pose of the given object such that it plausibly fits within the hands of the synthesized character. 
    }
    \label{fig:model_fig}
\end{figure*} 

\paragraph*{Full-Body Grasp Synthesis.}
This is a relatively recent line of work following the success of hand-object grasp synthesis. GOAL~\cite{taheri2021goal} synthesizes full-body motion for grasping a given object by first estimating the whole-body grasping pose for the object and then treating this pose as their goal for a motion-infilling module that interpolates the motion between a T-Pose and the goal pose.
SAGA~\cite{saga} also follows a similar strategy of motion infilling but uses markers to represent the body pose while also learning a contact map for the grasp for additional supervision.
Both these methods synthesize full-body motions \textit{until} the point of grasping.
In contrast, we synthesize the motion taking place \textit{after} the object is grasped (see Table~\ref{tab:in_out_com}).
This is a non-trivial and more challenging setup. Conditioning human and object motions based on the intended actions while also ensuring diversity in the generated motion sequences requires additionally learning their intent-based mutual interactions in an efficient and generalizable manner.

%
\section{Intent-Driven Full-Body Motion Generator}
\label{sec:Intent-Driven Full-Body Motion Generator}
We show the architecture of our intent-driven full-body motion generator model in Fig.~\ref{fig:model_fig}.
Given a human character's shape and initial 3D body pose, a rigid 3D object placed within their reach, and an intended action to perform with that object, our goal is to synthesize a full-body motion sequence of the character performing the intended action with the object.
We pose this problem as synthesizing the full-body motion sequence conditioned on the given object and a textual instruction label indicating the intent.
We solve this problem through four modules.
First, we encode the input instruction labels consisting of the type of the object and the associated action using our Condition Encoder. 
We also input the subject's body shape parameters into our Condition Encoder.
We use this encoding as a conditioning signal for all the modules.
A key characteristic of our problem is that the arms are the primary movers during human-object interactions.
Therefore, we use a pair of decoupled conditional variational autoregressor networks to synthesize the arm movements and the rest of the body movements separately, using an Arm Synthesis Module and a Body Synthesis Module, respectively.
Lastly, we use an Object Optimizer Module to optimize the $6$-DoF pose of the given object such that it fits plausibly within the hands of the synthesized character.

\subsection{3D Human Body and Object Representation}
We represent the human mesh using the SMPL-X~\cite{SMPL-X:2019} parametric body model. SMPL-X parametrizes the full human body along with the hands and the face as a differentiable function $SMPLX(\beta, \mathbf{r}, \Psi, \mathbf{t})$, consisting of body shape parameters $\mathbf{\beta} \in \mathbb{R}^{10}$, the root translation $\mathbf{t}\in \mathbb{R}^{3}$, the axis-angle rotations for the body joints $\mathbf{r} \in \mathbb{R}^{J\times3}$ $(J=55)$, and the face expression parameters $ \Psi \in \mathbb{R}^{10}$.
It maps the parameters to a body mesh with $10,475$ vertices.
To improve the stability and the convergence characteristics of our model, we use the 6D continuous representations~\cite{Zhou_2019_CVPR} $\mathbf{\theta} \in \mathbb{R}^{J\times6}$ to represent body joint rotations. 
We downsample all the objects in the dataset to $300$ vertices for faster optimization. 
The object's $6$-DOF pose is represented using a rotation matrix $\mathbf{R} \in \mathbb{R}^{9}$ and a translation vector $\mathbf{T}\in \mathbb{R}^{3}$.

\subsection{Model Design}
\label{sec:model_design}
We now discuss each of our modules in detail.
Our synthesis pipeline assumes that the character interacts with only one object at a time. 
Interactions can be either one-handed or both-handed, depending on the type of action and the object.

\subsubsection{Condition Encoder}
\label{CondEnc}
We input the object's category label using a one-hot vector $\mathbf{w_{o}}\in \mathbb{R}^{51}$. 
To represent the intended action information, we pass the intended action label, given as an English word, through the pre-trained CLIP~\cite{clip} model and use the embeddings $\mathbf{w_{a}}\in \mathbb{R}^{512}$ that it outputs.
The idea behind encoding the action labels with a pre-trained text encoder is the general relevance between the action semantics and the corresponding body movements. For example, actions such as ``drink'' and ``pour'' typically invoke similar arm movements and are semantically close. In contrast, other actions, such as ``inspect'' and ``pass'', invoke different body movements and are semantically different. 
Therefore, their embeddings, given by a large-scale language model such as CLIP, provide a regularized, semantics-based distribution of the intended actions and stabilizes further processing.  

We concatenate $\mathbf{w_{o}}$ and $\mathbf{w_{a}}$ with the body shape parameters $(\mathbf{\beta}\in \mathbb{R}^{10})$ and pass them into our Condition Encoder $\mathbf{q_c}$. Our Condition Encoder uses a series of MLPs to encode these input signals and projects them onto an encoded feature vector $\mathbf{\phi} \in \mathbb{R}^{400}$ as
\begin{equation}
    \mathbf{\phi} = \mathbf{q_c}(\mathbf{w_{o}}, \mathbf{w_{a}}, \mathbf{\beta}). 
\end{equation}

\subsubsection{Arm Synthesis Module}
Our Arm Synthesis Module is a conditional variational autoregressor that synthesizes the arm movements, conditioned on our condition encoder output $\phi$ and the previous $k$ frames of synthesized arm poses along with the 3D object positions.
The encoder of this module,  $\mathbf{q_a}$, takes in the tuple $\mathbf{q_a}^{in} = \braces{\mathbf{\phi}, \mathbf{\theta^a_{t-k:t-1}}, \mathbf{T_{t-k:t-1}}, \mathbf{R_{t-k:t-1}}}$, where $\mathbf{\theta^a_{t-k:t-1}}$ are the rotations for the arm joints synthesized by the past $k$ frames, and $\mathbf{T_{t-k:t-1}}$, $\mathbf{R_{t-k:t-1}}$ are the translation and rotation parameters of the object for the past $k$ frames.
During training, $\mathbf{q_a}$ uses a series of MLPs on the input and maps them to the parameters of a latent normal distribution, $\mathbf{\mu_a, \sigma_a} \in \mathbf{R}^{32}$.
The decoder, $\mathbf{\hat{q}_a}$, samples $\mathbf{z_a} \in \mathbb{R}^{32}$ from the latent distribution and uses the previous pose information $(\mathbf{q_a^{in}})$ to synthesize the arm pose for the current frame $(\mathbf{\hat{\theta}^a_t})$ through a series of MLPs with skip connections as
\begin{equation}
\mathbf{\hat{\theta}^a_t} = \mathbf{\hat{q}_a}(\mathbf{z_a}, \mathbf{q_a^{in}}).  
\end{equation}

\subsubsection{Body Synthesis Module}
Similar to the Arm Synthesis Module, the Body Synthesis Module is a   variational autoregressor.
We use the term `body' to denote the rest of the body parts apart from the arms. It includes the head, the torso, the hips, and the legs.
We also note that the movements of all these parts are correlated when performing a full-body action. 
For example, to drink from a cup, one has to tilt their head back when bringing the cup to their mouth.
To model such fine-grained correlations, we first compute a self-attention mapping between all the joints in each pose as
\begin{equation}
    \mathbf{\theta^{pe}_k} = \bracks{Attn(\mathbf{Q}, \mathbf{K}, \mathbf{V})}_\mathbf{k},
\end{equation} 
where the query $\mathbf{Q}$ is a joint position and the key-value pair $\parens{\mathbf{K}, \mathbf{V}}$ are information of all other joints provided as $J$ sinusoidal positional encodings for each of the $k$ frames. 
The encoder of the module,  $\mathbf{q_b}$, takes in the tuple $\mathbf{q_b}^{in} = \braces{\mathbf{\phi}, \mathbf{\hat{\theta}^a_{t-k:t-1}}, \mathbf{\theta^{pe}_{t-k:t-1}}, \mathbf{T_{t-k:t-1}}, \mathbf{R_{t-k:t-1}}}$.
The structure of $\mathbf{q_b}$ is similar to that of the Arm Synthesis Module encoder $\mathbf{q_a}$, and it maps the input $\mathbf{q_b^{in}}$ to the parameters of a latent normal distribution, $\mathbf{\mu_b, \sigma_b} \in \mathbf{R}^{100}$.
The decoder, $\mathbf{\hat{q}_b}$, samples $\mathbf{z_b} \in \mathbb{R}^{100}$ from the latent distribution and 
outputs the rest of the body poses as
\begin{equation}
\mathbf{\hat{\theta}^b_t} = \mathbf{\hat{q}_b}(\mathbf{z_b}, \mathbf{q_b^{in}}).  
\end{equation}
We then concatenate $\mathbf{\hat{\theta}^a_t}$ and $\mathbf{\hat{\theta}^b_t}$ to obtain the full-body pose $\mathbf{\hat{\theta}_t}$ at time $t$.
We pass $\mathbf{\hat{\theta}_t}$ to our Object Optimizer Module, along with the last predicted object position, to generate the object position for the current frame.
\subsubsection{Object Optimizer Module} \label{sec:optimization}
We have so far focused only on synthesizing the body poses for a given instruction. 
For a complete synthesis, we also need to estimate the corresponding $6$-DoF positions of the object.
Although fine-grained object synthesis is not the primary goal of our work, we aim to produce plausible object trajectories faithful to the synthesized full-body motion.
To this end, our core assumptions are that \textit{(a)} at the moment of grasping in the initial frame, the object is at rest in an upright position and \textit{(b)} inter-vertex distances between the vertices of the object and the hand remain constant throughout our intent-driven motion synthesis.

With these assumptions, we optimize the object's rotation $\mathbf{R}$, translation $\mathbf{T}$, as well as the pose parameters of the hand, $\mathbf{P}^h$, in the SMPL-X parameter space.

We first compute the matrix of Euclidean distances $\mathbf{D} \in \mathbb{R}^{N \times M}$ between the vertices on the hand, $\mathbf{V}^h \in \mathbb{R}^N$ and those on the surface of the object, $\mathbf{V}^o \in \mathbb{R}^M$ for the initial frame.
We can retrieve the hand vertices using the SMPL-X parameterization,
\begin{equation}
    \mathbf{V}^h = SMPLX(\mathbf{P}^h).
\end{equation}

For each subsequent frame, we then minimize the objective:
\begin{equation}
    \mathbf{R^*}, \mathbf{T^*}, \mathbf{P}^{h*} = \min_{\mathbf{R, T, P^h}} (\lambda_d E_d + \lambda_c E_c + \lambda_r E_r)
    \label{eq:ls_fit_1}
\end{equation}

\begin{figure}[t]
\centering
    \includegraphics[width=0.9\linewidth]{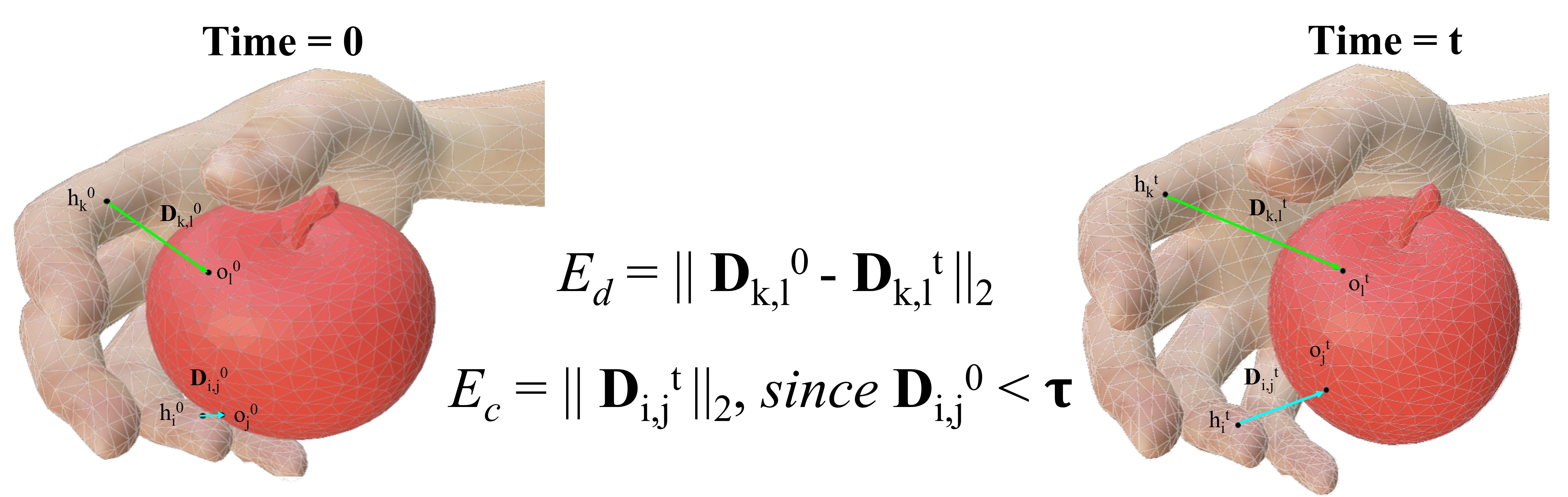}
    \caption{\textbf{Our Hand-Object Setup.} We design the energy term $E_d$ to enforce that the distances between the hand and the object vertices remain constant throughout the synthesis. Through the hand-object contact term $E_c$, we also enforce that the points in contact in the first frame remain in contact during the synthesis.}
    \label{fig:optimization_eqns}
\end{figure}

We use an energy term, $E_d$, to enforce the same inter-vertex distances between the hand and the object vertices in all the subsequent frames as in the first frame, as
\begin{equation}
    E_d (\mathbf{R, T, P^h}) = \norm{\operatorname{dist}(\mathbf{V^{h}}, \mathbf{R V^o} + \mathbf{T})  - \mathbf{D}}_2,
    \label{eq:ls_fit_2}
\end{equation}
However, this term alone does not guarantee that the object is in contact with the hand in subsequent frames because, in practice, the hand joints do not converge to plausible poses using $E_d$.
We address this issue by introducing the contact term $E_c$, which forces the distance between the in-contact vertex pairs of the first frame to be zero, as
\begin{equation}
    E_c(\mathbf{P^h}) = \norm{\delta . \operatorname{dist}(\mathbf{V^h}, \mathbf{R V^o} + \mathbf{T})}_2. 
\end{equation}
Here, $\delta(\cdot,\cdot)$ is a contact indicator function for the elements of the distance matrix for which the distance is less than a threshold: $\delta(i,j)) = 1$,  if $\mathbf{D}_{i,j} < \tau$ and $0$ otherwise, as we show in Fig.~\ref{fig:optimization_eqns}.

\begin{figure}[t]
\centering
    \includegraphics[width=0.9\linewidth]{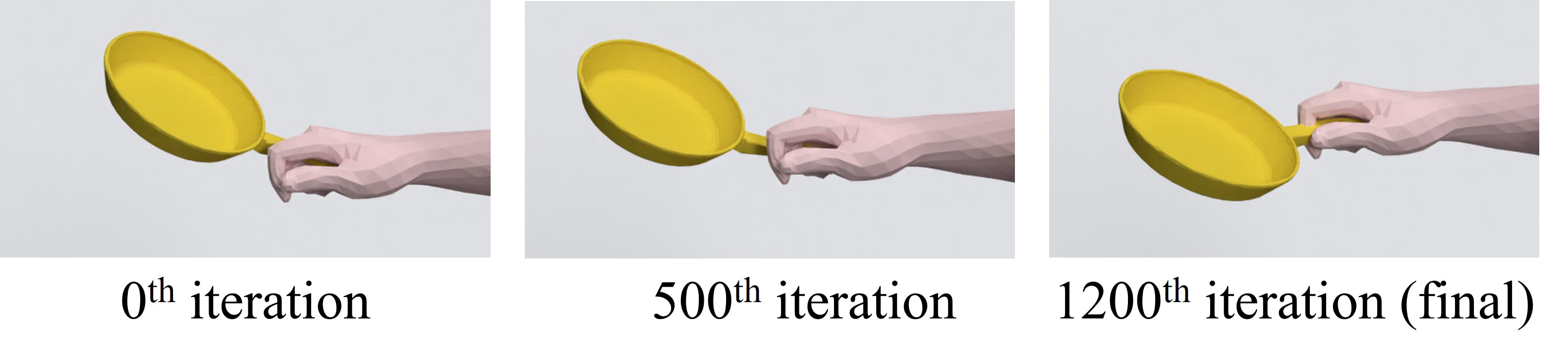}
    \caption{\textbf{Object Position Optimization}. We optimize the $6$-DoF pose of the object such that it fits plausibly within the hands of the virtual character. We show three snapshots of such fitting after the $0^{th}$, $500^{th}$ and the $1200^{th}$ iteration of our optimization.}
    \label{fig:optimization_iterations}
\end{figure}

Finally, $E_r$ consists of L2 regularizers to ensure that the object and hand poses do not deviate significantly from the previous frame and thus enforce temporal consistency, as
\begin{equation}
    E_r(\mathbf{R, T, P^h}) = \norm{\Delta{ \mathbf{R}} + \Delta{ \mathbf{T}} + \Delta{ \mathbf{P^h}}}_2,
\end{equation}
where $\Delta$ signifies the difference in values between the current and the previous frame.
We initialize the hand poses using a state-of-the-art grasp estimator proposed in~\cite{taheri2021goal}. 
The optimization routine iteratively corrects the initial estimates of the finger movements while placing the object within the person's hands.
Fig.~\ref{fig:optimization_iterations} illustrates the optimization routine.

\section{Implementation}
This section describes our training and inference routines and the implementation details for our generator network.

\paragraph*{Training and Inference Routines.}
To maintain a fixed number of input frames for computational stability, to reduce the parameter load and associated training overheads, and to avoid overfitting to redundant frames, we represent our ground-truth motion sequences using $T=15$ frames, taken at a sampling rate of $8$-$10$ fps.
These $15$ frames act as the key frames determining the motion sequence. 

The encoders and the decoders inside our four modules use fully-connected layers with skip connections, LeakyReLU activations, and batch normalization~\cite{agarap2018deep, NEURIPS2018_36072923}.
We use $k=4$ past frames (optimized through experiments) to synthesize the subsequent time steps.
We train our autoregressor based Arm Synthesis and Body Synthesis Modules to minimize the KL divergence loss:
\begin{align*}
    \mathcal{L}_{KL} =& D_{KL}\bigl(\mathbf{q}_{a}\parens{\mathbf{z_a}|\mathbf{x_{t-k:t-1}}, \mathbf{\phi}} || \mathcal{N}\parens{0, I}\bigr)  \\
    &+ D_{KL}\bigl(\mathbf{q}_{b}\parens{\mathbf{z_b}|\mathbf{x_{t-k:t-1}}, \mathbf{\phi}} || \mathcal{N}\parens{0, I}\bigr). 
    \numberthis\label{eq:kldiv_loss}
\end{align*}
We compute the pose and the velocity reconstruction loss between the ground-truth rotations $\mathbf{\theta}$ and the predicted rotations $\mathbf{\hat{\theta}}$ as
\begin{equation}
    \mathcal{L}_{rec} = \norm{\mathbf{\theta} - \mathbf{\hat{\theta}}}_1 + \norm{\Delta{\mathbf{\theta}} - \Delta{\mathbf{\hat{\theta}}}}_1. 
    \label{eq:rec_loss}
\end{equation}
We train our model on the following weighted sum of these losses:
\begin{equation}
    \mathcal{L} = \lambda_{KL}\mathcal{L}_{KL} + \lambda_p \mathcal{L}_{rec},
    \label{eq:total_loss}
\end{equation}
where $\lambda_{KL}$ and $\lambda_p$ are the weight parameters.
We can then use the regressed body motion parameters $\mathbf{\hat{p}}$ to optimize the $6$-DoF object positions at every time step.

During inference, we synthesize motions for novel intent-object pairs and novel body shape parameters. 
We input an initial body pose, a 3D object placed within reach of the character and an intended action to be performed with the object, and autoregressively synthesize the intent-based full-body motion sequence.

\paragraph*{Implementation Details.}\label{sec:implementation_details}
We train our model for $1$,$600$ epochs using the Adam Optimizer~\cite{adam} with a base learning rate of $5 \times 10^{-4}$, and a batch size of $64$, which takes roughly four hours on an NVIDIA A100-PCIE-40GB GPU.
We decay the learning rate (LR) using a Reduce-on-plateau LR scheduler 
with a patience of $3$ epochs and a decay rate of $0.999$. 
We set $\lambda_{KL}=0.001$,  $\lambda_p=\lambda_d=1.0$ and $\lambda_c=\lambda_r=0.005$. 
During inference, synthesizing the full-body poses and the corresponding object positions for a motion sequence of $15$ frames take approximately $1$-$1.5$ minutes.
Finally, we perform a linear interpolation on our generated frames to up-sample the motion to $30$ frames per sequence for cleaner visualization.
We have implemented our network, training, and inference using the PyTorch framework~\cite{pytorch}.
\section{Experiments and Results} 
\label{sec:experiments}
This section reports the results of our experimental evaluation, including the dataset, the evaluation metrics we use, and our ablation studies. 
Since there are no existing methods for generating full-body human-object interactions, we use existing methods that generate full-body poses based only on action labels as our baselines.

\subsection{Dataset}
\label{sec:dataset}
We use the GRAB dataset~\cite{GRAB:2020} consisting of whole-body grasping sequences performed by ten different subjects. 
The subjects interact with $51$ different objects via four basic intents, ``use'', ``pass'', ``lift'', and ``offhand''.
``Use'' further has a sub-category of $26$ different actions depicting plausible intent-object interactions such as drinking or pouring from a cup to taking a picture with or browsing a camera.
Following the split of \cite{diller2022charposes}, we take subject `S1' for validation, `S10' for testing, and the remaining subjects `S2' through `S9' for training.
This data split ensures that we test on novel subjects with different body shapes and our inference contains novel (intent-object) pairs such as offhanding a water bottle, which is not present in our training set.
We discard the ``lift'' intention from our setting as the motions depicting lifting an object were inconsistent in the dataset.
Thus our train, validation, and test splits respectively consist of $789$, $157$, and $115$ sequences.

\subsection{Baselines}
We compare our results with ACTOR~\cite{petrovich21actor}, Action2Motion~\cite{guo2020action2motion} and TEMOS~\cite{petrovich22temos}.
Since these methods were not originally trained on the GRAB dataset, we re-train them for our setting. 
We re-train ACTOR and the Action2Motion methods for $1$,$600$ epochs (the same number of epochs we train our model for, see Sec.~\ref{sec:implementation_details}) conditioned only on the action labels with no object information.
For comparison with TEMOS, we create sentences of the form ``A person $<action>$ the $<object>$'' (\textit{e.g.}, ``a person eats the apple'') to use as input sentences, and re-train the TEMOS model for $1$,$600$ epochs as well.
We apply our Object Optimizer Module for all three motion synthesis methods to generate the object positions for visual comparison.

\begin{table*}[t]
    \centering
     
    \begin{tabular}{lcccccc}
    \toprule
    Method & MPJPE ($\downarrow$) & AVE ($\downarrow$) & FID ($\downarrow$) & Accuracy ($\uparrow$) & Diversity ($\rightarrow$) & Multimodality ($\rightarrow$)   \\
    \midrule
    Real Motions (GT) & - & - & - & $0.97 \pm 0.001$ & $1.15 \pm 0.015$ & $0.30 \pm 0.010$  \\
    ACTOR & $0.09 \pm 0.005$ & $8.05 \pm 0.002$ & $0.67 \pm 0.002$ & $0.78 \pm 0.010$ & $1.06 \pm 0.015$ & $0.19 \pm 0.010$  \\
    Action2Motion & $0.11 \pm 0.003$ & $8.26 \pm 0.002$ & $1.08 \pm 0.002$ & $0.69\pm 0.011$ & $1.10 \pm 0.010$ & $0.22 \pm 0.010$  \\
    TEMOS & $0.10 \pm 0.005$ & $9.98 \pm 0.001$ & $1.21  \pm 0.004$ & $0.23 \pm 0.010$ & $0.83  \pm 0.010$ & $0.09  \pm 0.010$  \\
    \midrule
    Ablation 1 & $0.05 \pm 0.002$ & $4.41  \pm 0.002$ & $0.39 \pm 0.002$ & $0.78 \pm 0.012$ & $1.06  \pm 0.015$ & $0.21  \pm 0.010$ \\
    Ablation 2 & $0.04 \pm 0.005$ & $4.77 \pm 0.002$ & $0.38 \pm 0.002$ & $0.82 \pm 0.010$ & $1.10  \pm 0.020$ & $0.24  \pm 0.020$\\
    Ablation 3 & $0.05 \pm 0.005$ & $5.41 \pm 0.002$ & $0.42 \pm 0.002$ & $0.82  \pm 0.010$ & $1.08  \pm 0.010$ & $0.25  \pm 0.010$ \\
    Ours & $\mathbf{0.03} \pm \mathbf{0.005}$ & $\mathbf{3.82} \pm \mathbf{0.004}$ & $\mathbf{0.27} \pm \mathbf{0.002}$ & $\mathbf{0.87} \pm \mathbf{0.011}$ & $\mathbf{1.11} \pm \mathbf{0.015}$ & $\mathbf{0.28} \pm \mathbf{0.015}$  \\
    \bottomrule
    \end{tabular}

    \caption{ \textbf{Quantitative Evaluation.} We compare with other motion synthesis methods, namely ACTOR~\cite{petrovich21actor}, Action2Motion~\cite{guo2020action2motion} and TEMOS~\cite{petrovich22temos}, and three ablated versions of our model (Sec.~\ref{ablations}). We evaluate the methods on the MPJPE, AVE, FID, recognition accuracy, diversity, and multimodality metrics. ``$\downarrow$'' denotes lower values are better, ``$\uparrow$'' denotes higher values are better, and ``$\rightarrow$'' denotes values closer to the ground-truth are better.} 
    \label{tab:metrics}
    
\end{table*}

\subsection{Evaluation Metrics}
\label{quantitaive_metrics}
We evaluate our method using the Mean Per-Joint Positional Error (MPJPE), which measures the mean joint error over all time steps, and
the Average Variance Error (AVE)~\cite{Ghosh_2021_ICCV}, which
measures the variance error between the joint positions. 

We further evaluate the naturalness and the overall diversity of our generated motions using the Fr\'echet Inception Distance (FID)~\cite{heusel2017gans}, recognition accuracy, diversity, and multimodality. 
Following ACTOR~\cite{petrovich21actor} and Action2Motion~\cite{guo2020action2motion}, we train a standard RNN action recognition classifier on the GRAB dataset and use the final layer of the classifier as the motion feature extractor for calculating FID, diversity, and multimodality.

\begin{figure}[t]
\centering
    \includegraphics[width=0.9\linewidth]{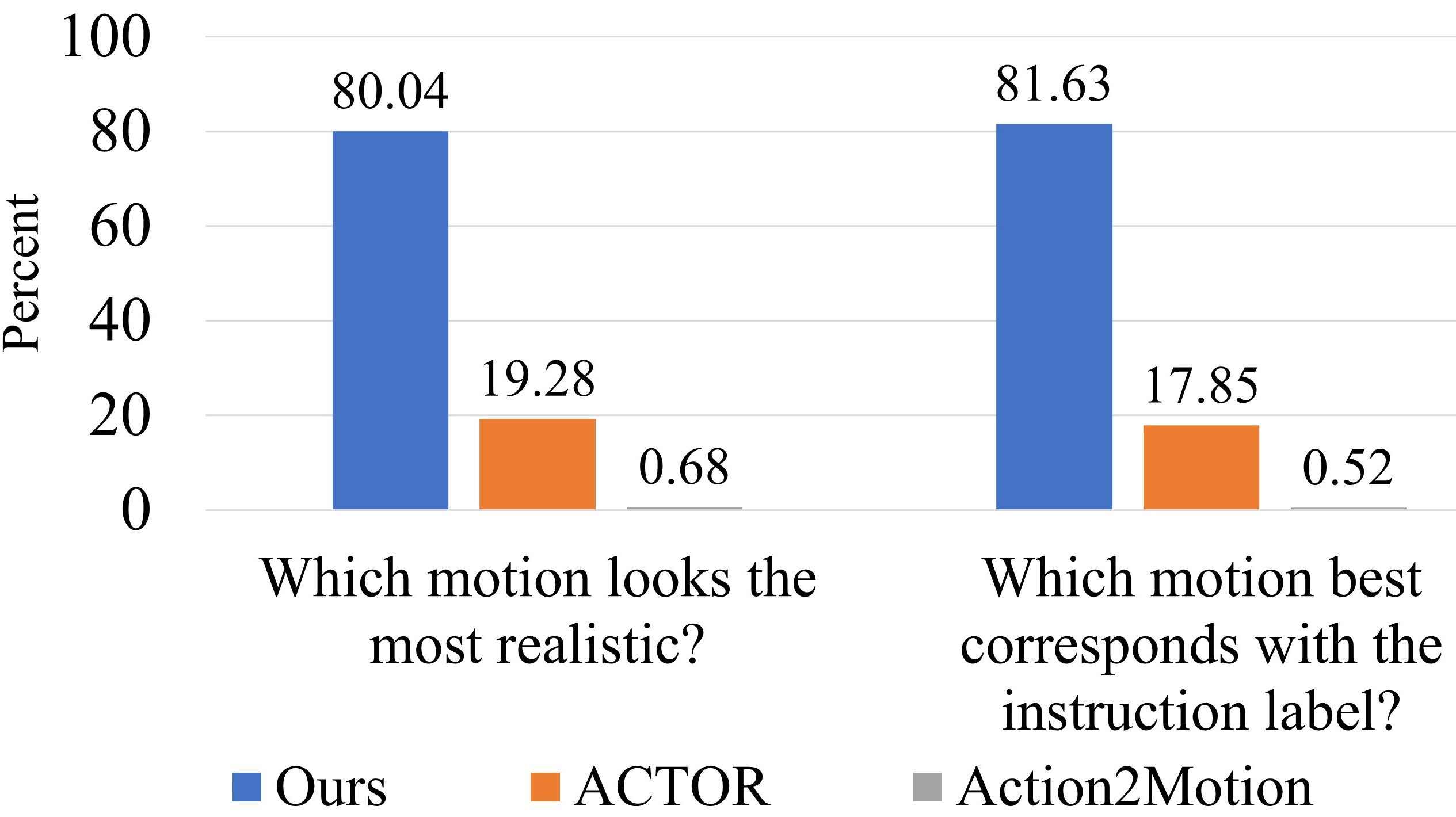}
    \caption{\textbf{Perceptual Study Evaluation.} We conduct a user study where participants answer two questions: ``Which animation looks more realistic?'' and ``which animation best corresponds with the input instruction label?''. We show them $30$ randomly sampled motion sequences synthesized by our method and the two baselines, ACTOR~\cite{petrovich21actor} and Action2Motion~\cite{guo2020action2motion}.
    We see our method is chosen more than $80\%$ times.}
    \label{fig:user_study}
\end{figure}

\begin{figure*}[t]
\centering
    \includegraphics[width=0.85\linewidth]{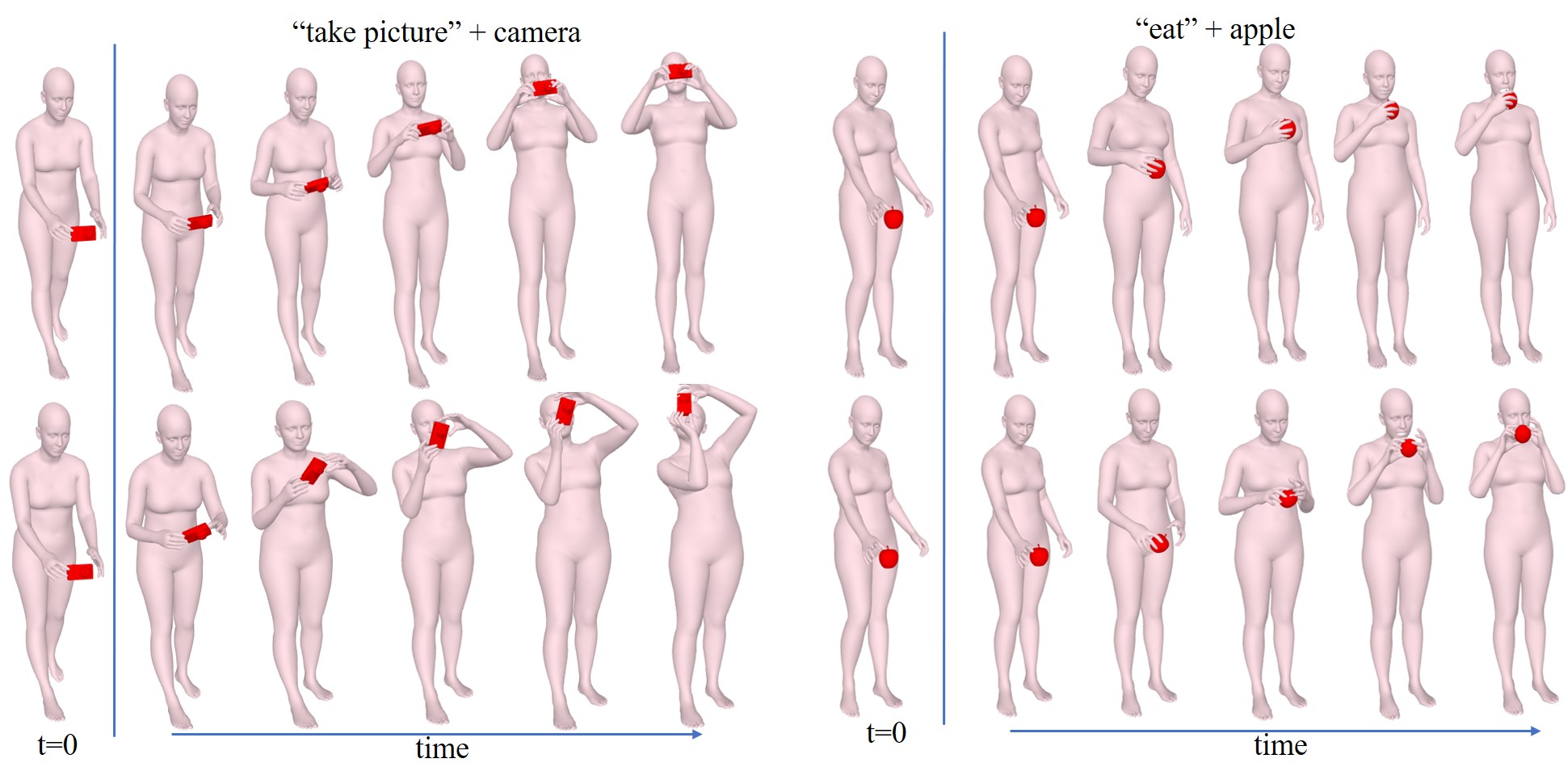}
    \caption{\textbf{Qualitative Results Showing Diversity in the Synthesized Motions.} The two rows depict two diverse motion sequences generated by our model. Our method can generate variations for the same instructions using either or both hands, along with plausible coordination of the head and the body. Please refer to the supplementary video for more results.}
    \label{fig:variations1}
\end{figure*}

\subsection{Ablation Studies}
\label{ablations}
We compare the performance of our model with the following ablated versions:
\begin{itemize}[leftmargin=*]
    \item \textbf{Ablation 1: Randomly initializing the input action labels with $\mathbf{512}$-d vectors:}
    To study how the CLIP model influences the conditioning of the synthesized motion, we conduct an ablation where we train our Condition Encoder with a $512$ dimensional randomly initialized vector for the input action labels instead of taking the CLIP embeddings. 
    \item \textbf{Ablation 2: Training the Body Synthesis module without using the self-attention mapping.} 
    In this ablation, we exclude our position-encoded multi-head self-attention from the input of the Body Synthesis module of our framework to see how it influences the quality of our motion. 
    \item \textbf{Ablation 3: Training the full body instead of decoupling to the Arm Synthesis and the Body Synthesis Modules.}
    We train the whole body movements in one module instead of separately synthesizing the arms and the rest of the body.
\end{itemize}

\subsection{Quantitative Evaluation} 

Table~\ref{tab:metrics} shows the MPJPE, AVE, FID, recognition accuracy, diversity, and multimodality on our test set compared to the three state-of-the-art methods of ACTOR~\cite{petrovich21actor}, Action2Motion~\cite{guo2020action2motion}, and TEMOS~\cite{petrovich22temos}.
We also include the ablated versions of our methods (Sec.~\ref{ablations}) in our evaluation. 
We repeat each experiment $20$ times as done in ACTOR~\cite{petrovich21actor}, and report a statistical interval with $95\%$ confidence.
Our method shows significant improvements in all the metrics compared to the existing methods and the ablated versions. 

\subsection{Perceptual Study} \label{perceptual_study}
To evaluate the visual quality of our motions, we conduct a perceptual study where we compare our results with ACTOR~\cite{petrovich21actor} and Action2Motion~\cite{guo2020action2motion}.
Except for TEMOS~\cite{petrovich22temos}, which would quickly settle on the mean pose, the other two methods generated plausible full-body motions after re-training. 
We, therefore, exclude TEMOS from the user study. 
We conduct our perceptual study in the following two sections.

\paragraph*{Comparison with Motion Synthesis Methods.}
%
In the first section, we displayed our results and the results from ACTOR and Action2Motion side-by-side in random order, along with the input instruction label.
We asked the participants to answer these questions for each sequence:
\textit{``Which motion looks the most realistic?'' } and \textit{``Which motion best corresponds with the input instruction label?''}.
We collected answers for $30$ such sequences from $75$ participants.
Fig.~\ref{fig:user_study} illustrates the results of the study. 
In $80\%$ responses, participants marked our method as the most realistic compared to ACTOR and Action2Motion.
Likewise, $81.6\%$ participants chose our method to have the best semantic fidelity with the instruction label.
Upon examining the cases for which the participants preferred ACTOR instead of us, we found that it performed better for a few actions, \textit{e.g.}, ``screwing'' the light bulb and ``toasting'' with the wineglass, where the motion does not need to have hand-to-eye or hand-to-mouth coordination.
These actions do not include significant variations within the dataset and are, therefore, easy to overfit.
\paragraph*{Comparison with Ground-Truth.}
While ACTOR and Action2Motion are the closest methods for our paradigm, they were not originally designed to synthesize motions conditioned on intents.
Therefore, to get an additional perspective on the performance of our method, we
asked the participants to compare our \textit{best} synthesis results with the ground-truth motions in the second section.
To establish an upper bound on our performance, we chose the $10$ best samples from various intent-object pairings to compare with the corresponding ground-truth motions.
Again, we displayed our synthesized and ground-truth motions side-by-side in random order.
This time, we kept an extra option: \textit{``cannot distinguish''}.
While our synthesized motions are, expectedly, less preferred than the ground-truth motions ($15.6\%$ vs. $36.9\%$),
$47.5\%$ of the responses rate our best syntheses as \textit{indistinguishable} in terms of realism. 
We also note that participants rated our synthesized motions as more realistic than the ground-truth motions when it involves actions such as ``eating'' an apple with one hand, which has abundant training samples.
On the other hand, our method encounters difficulties when synthesizing intents involving high-frequency wrist or finger movements such as ``shaking'' or ``squeezing''.
This is because our $\ell_1$ loss function (Eqn.~\eqref{eq:rec_loss}) tends to smooth out the high-frequency components from the motion sequence, and the GRAB dataset does not have sufficient samples of these actions to train them separately.

\subsection{Qualitative Evaluation}
We show full qualitative results in our supplementary video. 
When qualitatively compared with the ablated versions (Sec.~\ref{ablations}), we find that Ablation 1 (using random initialized vectors instead of CLIP) and Ablation 3 (training one module for the whole-body) fail to synthesize precise hand-mouth or hand-eye coordination for actions such as ``drinking'' and ``eating''.
Ablation 2 (without using self-attention mapping) lacks subtle body movements, such as tilting back the head or bending the knee to pick up an object, which improve the motion plausibility.
We further analyze our generated motions under the following headings:

\paragraph*{Diversity Analysis.}
As we noted earlier (Sec.~\ref{sec:intro}), generating diverse motion sequences for the same input instruction label is crucial for an immersive user experience.
Fig~\ref{fig:variations1} shows our result for two different sequences (left and right).
Sampling from the variational latent space allows us to synthesize diverse motion sequences. 
In Fig.~\ref{fig:variations1}, we show two different sequences: ``taking picture'' with a camera (left) and ``eating'' an apple (right). We show two variations of the same motions (upper and lower rows).
We note that the variations are diverse w.r.t how the head, arm, and torso are angled to use the object.
Our method benefits from operating in the full-body space and produces more natural results compared to na\"ively performing a fixed mapping from the global hand pose parameters to the end effectors of the remaining body.

\paragraph*{Synthesis of Both-Handed Interactions.}
Our method is the first to plausibly synthesize full-body motions for both-handed interactions.
We achieve this by decoupling the arm synthesis from the full-body synthesis in our generator design (Sec.~\ref{sec:model_design}).
The wrist and the elbow joints play a crucial role in tasks such as picking up an object with both hands or holding the object precisely.
Learning the arm motions in a separate latent space helps our generator focus more on such precise synthesis. 

\begin{figure}[t]
\centering
    \includegraphics[width=\linewidth]{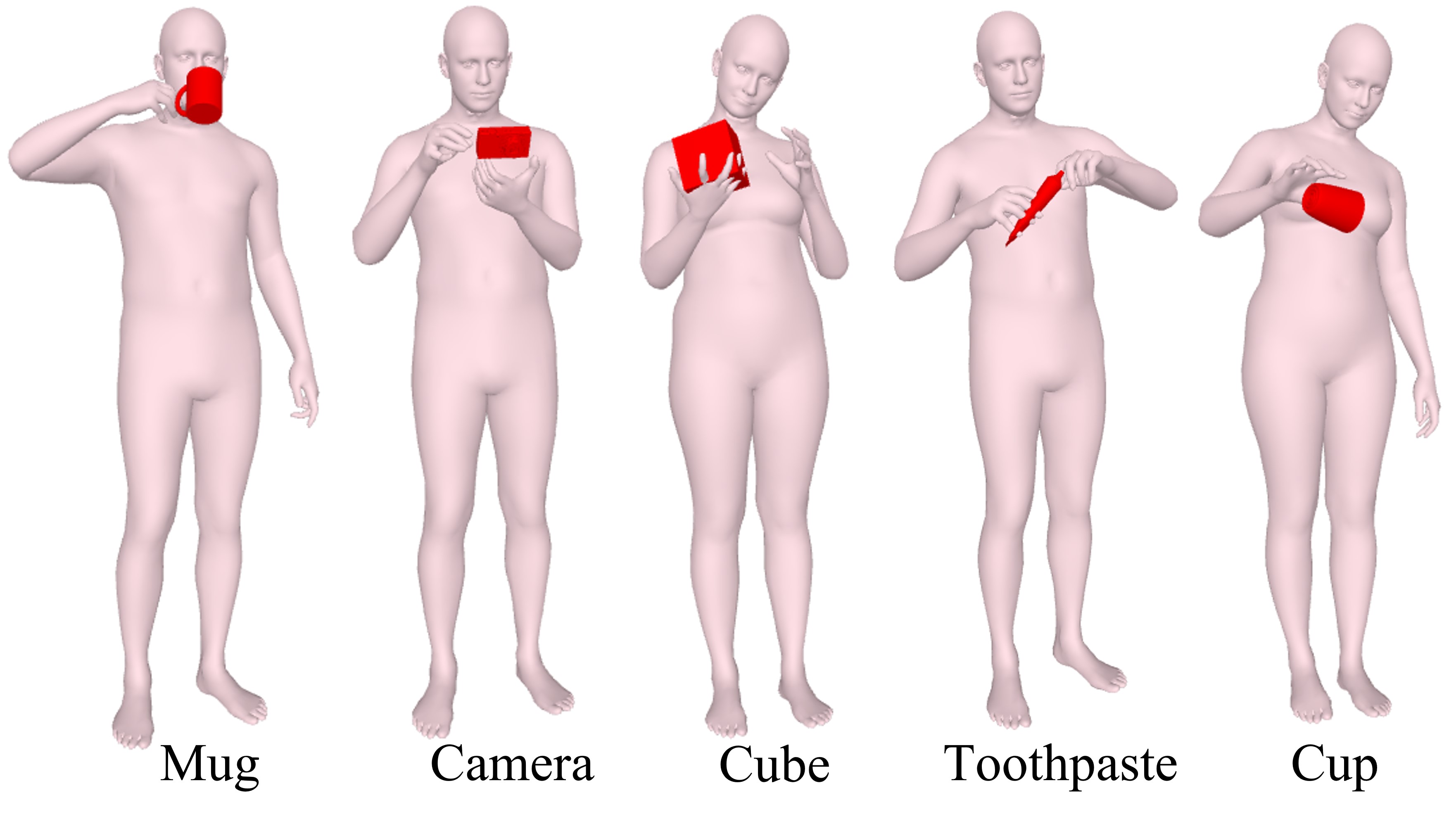}
    \caption{\textbf{Examples of Imprecise Contacts in the GRAB Dataset~\cite{GRAB:2020}.} We show five (ground-truth) frames where the body and the object are in contact.
    However, these contacts are not precise. The fingers do not touch the object when grasping the mug, the camera, or the cup.
    On the other hand, we see inter-penetration between the hand and the object for the cube and the toothpaste.} 
    \label{fig:GT_grasping}
\end{figure}

\begin{figure}[t]
\centering
    \includegraphics[width=\linewidth]{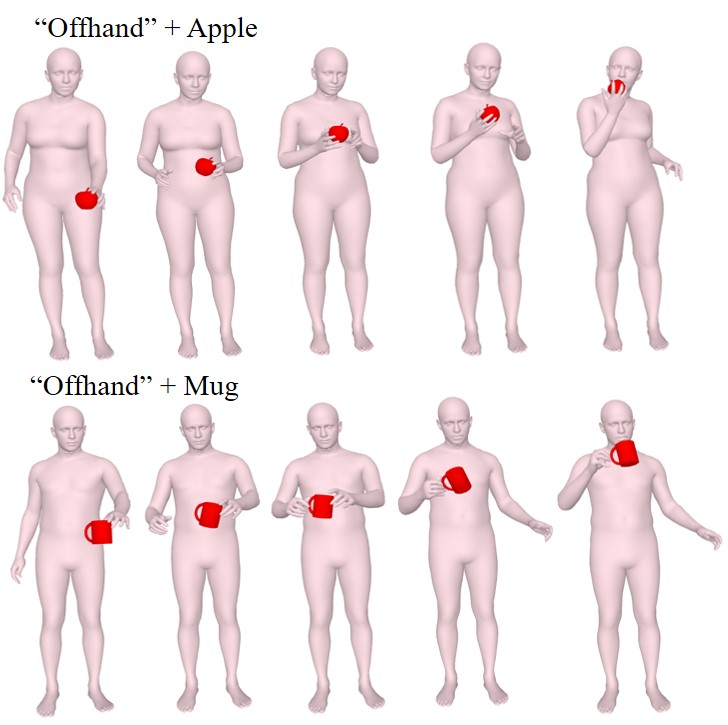}
    \caption{\textbf{Off-Handing.} We show two interactions of ``offhanding'' where the character passes the object from one hand to the other. Such interactions pose a unique optimization challenge when the object is switching hands.}
    \label{fig:offhand_intent}
\end{figure}

\paragraph*{Object Position Predictions for Off-Handing Interactions.}
In addition to both-handed interactions, we encounter sequences in the GRAB dataset where the character passes an object from one hand to the other.
It is non-trivial to optimize the accurate object positions when the object switches hands.
Here, we first compute the most likely frame at which the switching occurs and then transfer the optimized hand parameters to the other hand.
Fig:~\ref{fig:offhand_intent} shows two such off-handing interactions with two objects. 
\paragraph*{Plausibility of Head Motions.}
Similar to the motion of the fingers and the arms, the coordinated movement of the head and the hands also determines the synthesis quality.
While recent works like GOAL~\cite{taheri2021goal} explicitly account for the head direction vector during network training and optimization, we observe that our model learns visually plausible head orientations and hand-head coordination without any explicit supervision. This raises the question of whether explicit supervision is indeed necessary.

\section{Discussion and Limitations} 
Through quantitative evaluations and a perceptual study, we establish that our method synthesizes plausible motions of virtual characters performing intended actions with given objects.
While we can synthesize motions for various intents and objects, we observe failure cases for certain rare intents with high-frequency wrist motions, \textit{e.g.}, ``squeeze'', ``shake'' (see supplementary video).
Additionally, our Object Optimizer Module (Sec.~\ref{sec:optimization}) optimizes the fingers and the object positions based on an initial distance between them.
This assumption works well with most of the intents in the GRAB dataset, which involve static grasps.  
However, dynamic grasping, which involves hand slipping and relative motion between the object and the hands (such as ``rotating'' a cube and ``stretching'' an elastic band), is limited in our setting. We also note that the contacts between the body and the objects are not fully precise for all samples in the GRAB dataset, possibly due to the sparse marker-based motion capture.
In some sequences, the fingers do not touch the object while grasping, while in others, there are inter-penetrations between the hand and the object (Fig.~\ref{fig:GT_grasping}). Lastly, we do not address long-term motion synthesis (in the order of minutes) involving a sequence of actions performed with an object.

\paragraph*{Ethical Considerations.}
Our method does not support texture and fine appearance details and cannot be used to produce deceptive content. Our results are not photo-realistic by design and cannot be confused with real scenes.
However, combining our technique with a method supporting more realistic texture might raise ethical concerns in the future. 

\section{Conclusion and Future Work}

We presented the first full-body motion synthesis method for character-object interactions. 
Such a motion synthesis pipeline can become a useful, practical tool in applications requiring large-scale character animations.
We demonstrate that a decoupling approach that separately models the arms and the body motions using conditional variational autoregression leads to measurable perceptual improvements and advances the state-of-the-art on multiple quantitative evaluations.
We also synthesize interactions involving \textit{both} hands, including sequences where the object exchanges hands.

In the future, we intend to extend our model to synthesize dynamic grasps and full-body poses such that the virtual character can change the grasp within a sequence. 
We also plan to explore descriptive sentence embeddings for the interactions (\textit{e.g.}, ``a person passes the bowl using the right hand'') to generate more precise and controllable motions.

\paragraph*{Acknowledgements.}
This research was supported by the BMBF grant XAINES (01$|$W20005), the BMBF and ITEA grant AIToC (01$|$S20073B), the EU Horizon 2020 grant Carousel+ (101017779), and the ERC Consolidator Grant 4DRepLy (770784).

\printbibliography 

\end{document}